\pdfoutput=1

\documentclass[11pt]{article}
\usepackage[]{acl}
\usepackage{times}
\usepackage{latexsym}
\usepackage[T1]{fontenc}
\usepackage[T5]{fontenc}
\usepackage{CJK}
\usepackage{multirow}
\usepackage[utf8]{inputenc}
\usepackage{soul}
\usepackage{amssymb}
\usepackage{microtype}
\usepackage{graphicx}
\usepackage{booktabs}

\usepackage{pifont}
\title{
The Cross-lingual Conversation Summarization Challenge}

\author{{
 Yulong Chen\thanks{\ \ Equal Contribution.}$^{*\spadesuit \heartsuit}$\hspace{0.5mm}, 
 Ming Zhong$^{*\clubsuit}$\hspace{0.5mm},
 Xuefeng Bai$^{*\spadesuit \heartsuit}$\hspace{0.5mm},
 Naihao Deng$^{\ddagger}$\hspace{0.5mm},}\\
 {\large\bf{Jing Li$^{\diamondsuit}$\hspace{0.5mm},
 Xianchao Zhu$^{\diamondsuit}$\hspace{0.5mm}, 
 Yue Zhang$^{\heartsuit}$\hspace{0.5mm}}} \\
 $^\spadesuit$ Zhejiang University\\
 $^\heartsuit$ School of Engineering, Westlake University\\
 $^\clubsuit$ University of Illinois at Urbana-Champaign\\
 $^\ddagger$ University of Michigan, Ann Arbor \\
 $^\diamondsuit$ Sichuan Lan-bridge Information Technology Co., Ltd.\\
 \texttt{yulongchen1010@gmail.com}\quad\texttt{yue.zhang@wias.org.cn}
}
\date{}

\begin{document}
\maketitle
\begin{abstract}
We propose the shared task of cross-lingual conversation summarization, \emph{ConvSumX Challenge}, opening new avenues for researchers to investigate solutions that integrate conversation summarization and machine translation.
This task can be particularly useful due to the emergence of online meetings and conferences.
We construct a new benchmark, covering 2 real-world scenarios and 3 language directions, including a low-resource language.
We hope that \emph{ConvSumX} can motivate researches to go beyond English and break the barrier for non-English speakers to benefit from recent advances of conversation summarization.


\end{abstract}

\section{Task Overview}

\begin{figure*}
    \centering
    \includegraphics[width=\textwidth]{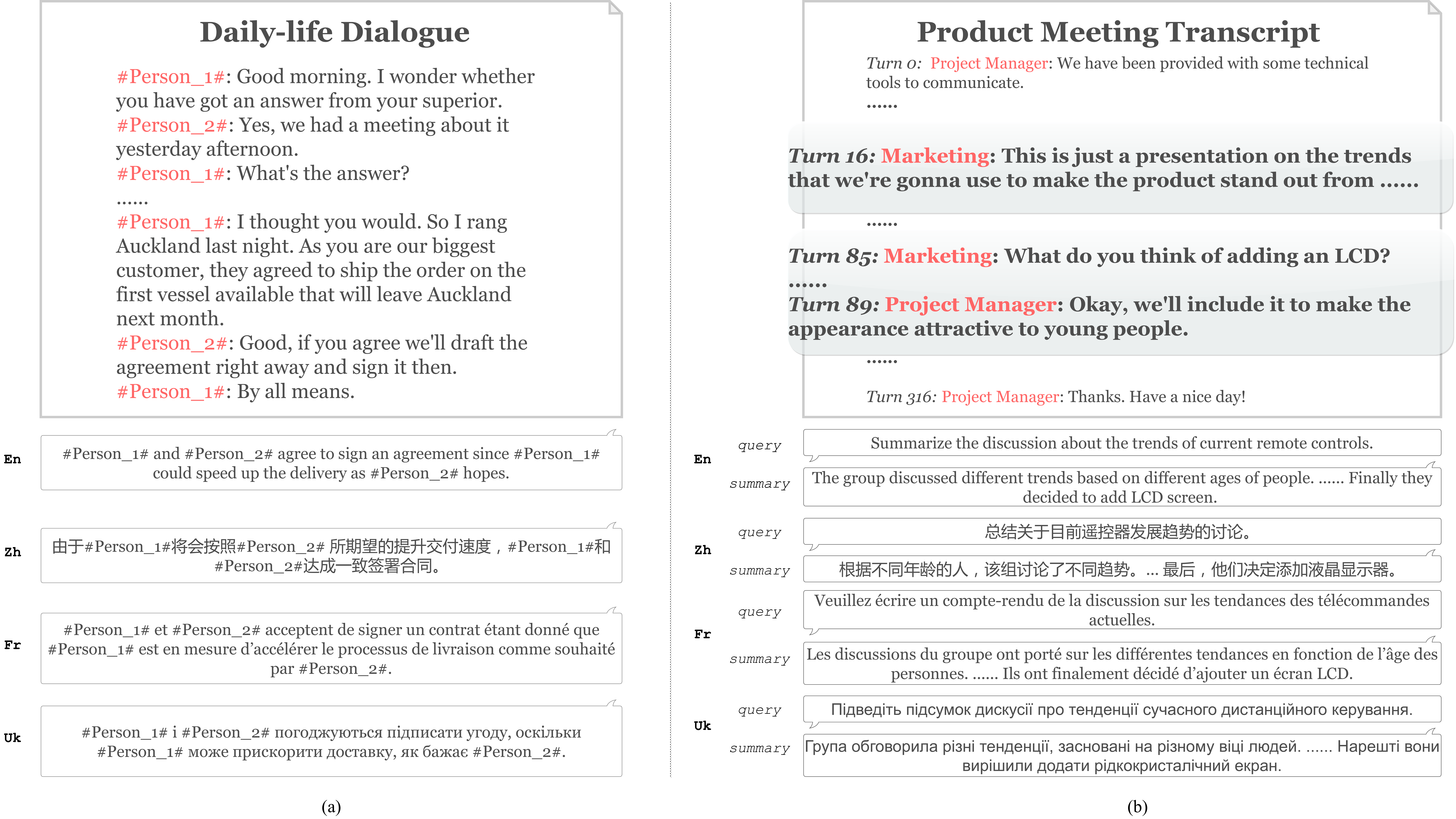}
    \caption{Examples of \emph{ConvSumX}.
    Given the conversation text (and the \emph{\texttt{query}}), the task is to generate a \emph{\texttt{summary}}.
    From top to bottom, the languages are English (\texttt{En}), Chinese (\texttt{Zh}), French (\texttt{Fr}) and Ukrainian (\texttt{Uk}).
    The English summaries (and queries) are from the original datasets.
    }
    \label{fig:qmsum}
\end{figure*}

The cross-lingual conversation summarization (\textit{ConvSumX}) task asks models to output a salient, concise, fluent and coherent summary in the target language (\emph{e.g.}, Chinese), given a conversation in the source language (\emph{e.g.}, English).
In particular, \textit{ConvSumX} contains 2 tracks: (1) daily dialogue summarization and; (2) query-based meeting minute. 
Each track covers 3 language directions: \texttt{English-to-Chinese} (\texttt{En2Zh}), \texttt{English-to-French} (\texttt{En2Fr}) and \texttt{English-to-Ukrainian} (\texttt{En2Uk}).
Figure~\ref{fig:qmsum} gives an example for each track in \emph{ConvSumX} respectively, where we show summaries (queries) in 4 languages (including English).
Both automatic and manual evaluations are used to measure the performance of submitted systems, while the evaluation is highly inclined to human evaluation focusing on cross-lingual consistency
(Section~\ref{evaluation}).


\section{Motivation}\label{motivation}
\label{sec:motivation}
Thanks to the availability of large-scale corpora~\cite{gliwa2019samsum,chen-etal-2021-dialogsum,zhong-etal-2021-qmsum}, research on conversation summarization has made great progress~\cite{zhong2021dialoglm,ni-etal-2021-summertime,ghazvininejad2021discourse,linother}.
However, existing corpora in this area focus on English while ignoring other languages~\cite{feng2021survey}.
Such English-dominated corpora lead to a barrier for non-English speakers to benefit from conversation summarization research, which becomes more severe in the era of epidemic, where international meetings are held online and participants communicate in English.

\emph{ConvSumX} integrates conversation summarization and machine translation, involving the language shift from one to another and stylistic shift from long spoken conversations to concise written monologues.
Ideally, using the \emph{first translate, then summarize} and vice versa pipelines
can solve the task.
However, besides the difficulties that have been discussed in monolingual conversation summarization~\cite{chen-etal-2021-dialogsum-challenge,DBLP:conf/acl/FengFQ0020}, pipelines methods suffer from problems caused by machine translation systems.
For example, for \emph{translation-first} pipelines, most existing machine translation systems show poor performance on conversation text~\cite{wang-etal-2021-autocorrect}. 
For \emph{summarization-first} systems, translating summaries without considering conversation context can lead to inconsistent translation, in particular for polysemous words.
Take $\texttt{C}^\texttt{En}$\texttt{2}$\texttt{S}^\texttt{Zh}$~\footnote{The setting means the input conversation text is in English, and the output summary is in Chinese.} for example. 
The summary ``\emph{Bob is going to the bank.}'', where ``\emph{bank}'' can be translated into ``\begin{CJK}{UTF8}{gbsn}岸边\end{CJK}'' (river bank) or ``\begin{CJK}{UTF8}{gbsn}银行\end{CJK}'' (financial bank), requires models to determine the proper translation by considering conversation context. 
Such issues can be also in end-to-end systems developed for cross-lingual news summarization and directly using those methods can lead to error propagation~\cite{zhu-etal-2019-ncls,xu-etal-2020-mixed,liang2022variational}.
Thus, more sophisticated designs that take care of conversation natures or data selection strategies that can make better use of silver data are in need.

To this end, \emph{ConvSumX Challenge} encourages researchers to investigate different solutions to cross-lingual conversation summarization.
First, from the perspective of downstream applications, \emph{ConvSumX} is useful for both business and personal uses.
Second, from the perspective of research, \emph{ConvSumX Challenge} looks for a general method that can deal with cross-lingual conversation summarization.
Although we present 3 typical target languages in this work, we hope that \emph{ConvSumX} can motivate researches to a broader range of target languages.
Third, from the perspective of social good, \emph{ConvSumX} aims to break the barrier of accessing information for non-English speakers and to make them benefit from the advance of conventional English-dominated conversation summarization technologies.

We hope that \emph{ConvSumX} can gain interests from both communities of text summarization and machine translation and also push the progress on related fields for other languages, including more low-resource languages.



\section{Task Description}
Formally, the task of \emph{ConvSumX Challenge} asks participants to provide a system that can output a summary in a target language given the input conversation text in a source language.

\subsection{Setting}
The \emph{ConvSumX Challenge} focuses on the low-resource/few-shot setting and cross-lingual/domain transfer technologies.
The low-resource/few-shot here is stated from the perspective of \emph{lacking large gold training data}.
The term \emph{gold data} refers to cross-lingual $\langle$\texttt{conversation-summary}$\rangle$ pairs that are annotated by experts.

\begin{table*}[t]
    \centering
    \small
    \begin{tabular}{l|cccccc}
    \toprule
        \textbf{Track} & \textbf{Data Source} & \textbf{Domain} & \textbf{Query} & \textbf{\# Conv.} & \textbf{\# Summ.} & \textbf{Train/Dev/Test} \\
        \midrule
        Track 1 & \textsc{DialogSum} & Daily-life Dialogue & \ding{55}& 131.0 &13.8& 12,460/500/500 \\
        \midrule
        \multirow{2}{*}{Track 2} &  \multirow{2}{*}{QMSum} & Product Meeting & \ding{51}& 6,007.7 & 70.5 & 690/145/151 \\
        & & Academic Meeting &\ding{51} & 13,317.3 & 52.7 & 259/54/56\\
        \bottomrule
    \end{tabular}    
    \caption{Statistics of \emph{ConvSumX}. 
    \# Conv. and \# Summ. are averaged lengths of conversations and summaries.
    }
    \label{tab:stats}
\end{table*}

The reasons are:
(1) gold parallel data are limited as the annotation is costly, in particular when source conversations involve domain expert knowledge (\emph{e.g.,} academic meeting). 
In contrast, machine translation data and monolingual summarization data are abundant and useful~\cite{perez-beltrachini-lapata-2021-models};
(2) we seek for a general solution that can be applied to not only the target languages as in this paper, but also other languages.
However, for practical consideration, we provide large silver data generated by multiple translation systems (\emph{cf. Section~\ref{annotation}}).
We encourage participants to make use of external resources to solve the task.

The above setting is widely adopted by existing cross-lingual summarization datasets in other domains, such as the first large-scale cross-lingual summarization corpus, NCLS dataset~\cite{zhu-etal-2019-ncls} and its succeeding works~\cite{xu-etal-2020-mixed,bai-etal-2021-cross,liang2022variational}.

\subsection{Tracks}
The \emph{ConvSumX Challenge} consists of 2 tracks, focusing on different scenarios, respectively.

\textbf{Track 1} focuses on cross-lingual summarization for real-life dialogues.
This track is in line with the INLG 2021 \emph{DialogSum Challenge}~\cite{chen-etal-2021-dialogsum-challenge} while we extend \emph{DialogSum} into a cross-lingual setting.
\emph{ConvSumX} can be particularly useful in scenarios such as travelling or studying abroad where summarizers can serve as personal assistants for non-English speakers.

\textbf{Track 2} focuses on cross-lingual meeting minutes.
Compared with daily conversations, meeting conversations are much longer and contain rich topic switches and more professional knowledge.
Generating cross-lingual meeting minutes can help non-English speakers to quickly access information of their interest, especially in the case where conferences are mostly held in English.
In particular, Track 2 asks a system to generate a summary in the target language, given an input meeting text in the source language and a query in the target language.
\subsection{Data}
\subsubsection{Data Selection}
The data of \emph{ConvSumX} are derived from two public English datasets, namely \textsc{DialogSum}~\citep{chen-etal-2021-dialogsum} and QMSum~\citep{zhong-etal-2021-qmsum}.
We choose these two datasets for: (1) their high-quality annotation and; (2) that they are suitable for real-life applications.
Table~\ref{tab:stats} shows the statistics.

\textbf{\textsc{DialogSum}} is a large-scale real-life scenarios dialogue summarization dataset, consisting of face-to-face spoken dialogues that cover a wide range of daily-life topics.
In particular, \textsc{DialogSum} provides multi-references for each dialogue in test set.
We ask annotators to first choose the best reference summary from multi-references and then annotate it into the target languages.

\textbf{{QMSum}} is a query-based meeting minute dataset, covering 3 domains, namely academic, product and committee.
We choose academic meeting and product discussion meeting for annotation as they are more in line with our motivation.

\subsubsection{Annotation}\label{annotation}
We invite native translators as our annotators and ask them to annotate summaries in dev and test sets of \textsc{DialogSum} and QMSum into 3 target languages including Chinese, French and Ukrainian.
Note that our data annotation is not the simple translation of summaries, instead, each annotation needs to take care of original English conversations to ensure that the annotated cross-lingual summary is consistent with the input (\emph{cf.} Section~\ref{sec:motivation}).

Besides manually annotated dev and test sets, following \citet{zhu-etal-2019-ncls}, we construct silver training data using machine translation.
In particular, we translate summaries in target languages using multiple engines, including Google translate~\footnote{\url{https://translate.google.com}}, NiuTrans~\footnote{\url{https://niutrans.com}} and LanMT~\footnote{\url{https://www.dtranx.com}}.
In addition, to provide resources for pipeline methods, we translate conversation texts using the same methods.

\subsection{Evaluation}\label{evaluation}
The evaluation of the \emph{ConvSumX Challenge} considers both automatic and manual evaluation metrics.

\subsubsection{Automatic Evaluation}
Following previous cross-lingual summarization work~\cite{zhu-etal-2019-ncls}, we use \textsc{Rouge} scores~\cite{lin2004rouge} for automatic evaluation.
\textsc{Rouge} scores evaluate the model performance by considering the overlap of $n$-grams in the system-generated summary against the reference summary.
Although recent works claim that \textsc{Rouge} fails to measure important information regarding factual consistency~\cite{DBLP:conf/iclr/ZhangKWWA20,10.1162/tacl_a_00373}, we choose \textsc{Rouge} because: (1) it directly reflects model's ability of obtaining salient information and; (2) it can be easily applied to multiple languages including low-resource language.

\subsubsection{Manual Evaluation}
As neural summarizers mostly contain factual errors that cannot be easily detected by automatic metrics~\cite{zhu-etal-2019-ncls,10.1162/tacl_a_00373} and translated words can be various~\cite{10.1162/tacl_a_00437}, automatic evaluation such as \textsc{Rouge} can be less accurate. Thus, our evaluation highly relies on manual evaluation. 
Given that the \emph{ConvSumX} integrates conversation summarization and machine translation, we adopt multiple human evaluation metrics from both tasks to better measure model performance.

In particular, standard summarization metrics include: Fluency, Consistency, Relevance and Coherence~\cite{kryscinski2019neural}; standard machine translation metrics include: Omission, Untranslation, Mistranslation, Addition and Terminology~\cite{mariana2014multidimensional}.
However, except for Fluency, summarization metrics evaluate generated summaries from the perspective of input documents in the same language while machine translation metrics evaluate translation from the perspective of source sentences (the English summary in our case).
There can be an evaluation inconsistency between these two tasks.
In addition, there is overlap between these two groups of metrics. For example, a mistranslated summary can be regarded as containing consistency errors.

To unify the aforementioned evaluation metrics and obtain fine-grained evaluations, we propose to evaluate system-generated summaries from following aspects against source conversation texts.

\textbf{Fluency} evaluates the quality of individual sentence, including the grammar, word order, etc.

\textbf{Coherence} evaluates the collective quality of all generated sentences.

\textbf{Relevance} evaluates the importance of information in the generated summary.

\textbf{Consistency} evaluates factual alignment of the generated summaries against the source conversation, including information that is not presented in the conversation, wrong causal relation, etc.

\textbf{Terminology} evaluates the use of language. For example, the generated word can be a right translation but is improper in certain domains (\emph{e.g., } academic meeting).

\textbf{Untranslation} judges whether the generated summaries contain untranslated words.

\textbf{Overall score} measures the overall quality for each summary.

For each metric above, we randomly extract 10\% generated summaries and ask annotators to give scores from $1$ to $5$.
The higher, the better. 

\section{Related Work}
\citet{zhu-etal-2019-ncls} propose the first large scale cross-lingual news summarization dataset, facilitating the study in this filed using neural network models.
\citet{bai-etal-2021-cross} construct an \texttt{English-to-German} news summarization dataset using the automatic method of \citet{zhu-etal-2019-ncls}.
\citet{perez-beltrachini-lapata-2021-models} construct a cross-lingual dataset based on Wikipedia, focusing on European languages.
In particular, \citet{perez-beltrachini-lapata-2021-models} use the lead paragraph in other languages aligned by Wikipedia interlanguage links, and the original document to construct $\langle$\texttt{document-summary}$\rangle$ pairs.
Similarly, \citet{ladhak-etal-2020-wikilingua} construct the WikiLingua dataset based on multi-lingual WikiHow.

Very recently, \citet{wang2022clidsum} and \citet{feng-etal-2022-msamsum} construct cross-lingual dialogue summarization datasets.
In particular, \citet{wang2022clidsum} manually translate summaries from SAMSum~\cite{gliwa2019samsum}, an online written dialogue summarization dataset, and 40k data in MediaSum~\cite{zhu-etal-2021-mediasum} into German and Chinese.
\citet{feng-etal-2022-msamsum} construct MSAMSum by automatically translating SAMSum into Chinese, French and Russian.
Compared with them, our work focuses on spoken conversation in multiple scenrios, and cover low-resource language (\texttt{Ukrainian}). 
In addition, we also focus on query-based meeting scenarios, which can be more useful in real-world applications.

\section{Conclusion}
We propose the \emph{ConvSumX Challenge} to address the task of cross-lingual conversation summarization, with the hope that \emph{ConvSumX} can encourage researchers to investigate various methods for conversation summarization beyond the English, in particular for low and mid-resource languages.

\section{Acknowledgement}
Yue Zhang is the corresponding author.
We thank Yang Liu, Da Yin and Jianyu Wang for insightful discussion and proofreading.
Our sincere appreciation goes to professional translators from Sichuan Lan-bridge Information Technology.

\section{Ethical Consideration}
\subsection{Copyright and License of Datasets}
The \emph{ConvSumX Challenge} uses cross-lingual $\langle$\texttt{conversation-summary}$\rangle$ pairs that are annotated on the top of two English conversation summarization datasets, namely \textsc{DialogSum} and QMSum, to evaluate models.
Both \textsc{DialogSum} and QMSum are free for academic use with the MIT license, which contains no limitation to use, modification or distribution.
We will also make our annotated data available for the academia.

\bibliography{anthology,custom}
\bibliographystyle{acl_natbib}




\end{document}